\documentclass[review]{elsarticle}

\usepackage{lineno,hyperref}
\usepackage{color,soul} 
\usepackage{amsmath}
\usepackage{graphics} 
\usepackage{multirow}
\usepackage[flushleft]{threeparttable}
\usepackage[export]{adjustbox}

\begin{document}

\begin{frontmatter}

\title{Improving forecasting accuracy of time series data using a new ARIMA-ANN hybrid method and empirical mode decomposition}


\author[metu]{\"{U}mit \c{C}avu\c{s} B\"{u}y\"{u}k\c{s}ahin\corref{mycorrespondingauthor}}
\cortext[mycorrespondingauthor]{Corresponding author}
\ead{umit.buyuksahin@ceng.metu.edu.tr}
\author[metu,mit]{\c{S}eyda Ertekin}
\ead{seyda@ceng.metu.edu.tr}
\address[metu]{Department of Computer Engineering, Middle East Technical University (METU)\\\"{U}niversiteler Mh., No:1, 06800, Ankara, Turkey }
\address[mit]{MIT Sloan School of Management, Massachusetts Institute of Technology\\77 Massachusetts Ave, Cambridge, MA 02139, USA}



\begin{abstract}

Many applications in different domains produce large amount of time series data. Making accurate forecasting is critical for many decision makers. Various time series forecasting methods exist which use linear and nonlinear models separately or combination of both. Studies show that combining of linear and nonlinear models can be effective to improve forecasting performance. However, some assumptions that those existing methods make, might restrict their performance in certain situations. We provide a new Autoregressive Integrated Moving Average (ARIMA)-Artificial Neural Network(ANN) hybrid method that work in a more general framework. Experimental results show that strategies for decomposing the original data and for combining linear and nonlinear models throughout the hybridization process are key factors in the forecasting performance of the methods. By using appropriate strategies, our hybrid method can be an effective way to improve forecasting accuracy
obtained by traditional hybrid methods and also either of the individual methods used separately.

\end{abstract}

\begin{keyword}
Time series forecasting, ANN, ARIMA, Hybrid methods, Empirical mode decomposition (EMD)
\end{keyword}

\end{frontmatter}


\section{Introduction and related work}
Time series include data points listed in time order. It is generally a sequence of discrete-time data which consists of points equally spaced in time. In time series forecasting, we try to predict the future points by analyzing observed points in the series.  It has been widely used in various applications of science, engineering, and business fields. However,  time series data might show different characteristics and show increasing or decreasing trends. Some time series has seasonal trends in which variations are specific to a particular time range, e.g coat and boot sales increase in winter season while decrease in summer season. On the other hand, some time series are not seasonal, such as stock market data. Moreover, time series data might show different level of volatility. While USD/EUR (United States dollar/Euro) exchange rate shows high volatility, growth of an animal, plant, or human being show a linear change. Over past several decades, a considerable effort has been devoted to develop and improve time series forecasting models~\cite{ph.d_sas_2018}. In the literature, various forecasting methods have been proposed which  use  linear and nonlinear models separately or combination of both. In this paper, we propose a hybrid algorithm of linear and nonlinear methods where we choose Autoregressive Integrated Moving Average (ARIMA) as a linear method and  Artificial Neural Networks (ANNs) as a nonlinear method.

 ARIMA is widely used linear time series forecasting method that is used in numerous applications including finance~\cite{contreras_arima_2002}, engineering~\cite{ gonzalez-romera_monthly_2006-1}, social sciences~\cite{k_p_modelling_2016}, and agriculture~\cite{wang_improving_2015}.  ARIMA models are integration of Autoregressive models (AR) and Moving Average models (MA). The model building process of ARIMA depends on Box-Jenkins methodology \cite{box_forecasting_2008} where it provides a step-by-step procedure for AR and MA analysis. ARIMA models give good accuracy in forecasting relatively stationary time series data. However it makes a strong assumption that the future data values are linearly dependent on the current and past data values. In addition, it also assumes that there is no missing data in the given time series~\cite{ediger_arima_2007}. 
 However, many real world time series data presents complex nonlinear patterns which might not be modeled by ARIMA effectively.  
 
For the nonlinear time series modeling, Artificial Neural Networks (ANNs) are one of the most widely used algorithms \cite{lapedes_nonlinear_1987} in many fields, such as finance ~\cite{chen_using_2009}, energy~\cite{singhal_electricity_2011}, hydrology \cite{ardabili_computational_2018}, and network communications \cite{chang_novel_2009}. ANNs have several advantages over ARIMA and other forecasting models. Firstly, ANNs are capable of fitting  a complex nonlinear function, therefore it does not need to make data stationary as ARIMA does. This ability helps ANNs to approximate any continuous measurable function with arbitrarily desired accuracy \cite{cybenko_approximation_1992}, \cite{hornik_multilayer_1989}. Moreover, ANNs are adaptively data-driven in nature  which means ANN models can be adaptively formed based on the features of time series data \cite{zhang_zhang_2003}. 

In the literature, there are studies which show the success of linear and nonlinear methods over each other. For example, \cite{foster_neural_1992, brace_comparison_1991, aras_new_2016} report that statistical and linear models give better results than ANNs. On the other hand, \cite{denton_how_good_1995, hann_much_1996, callen_neural_1996} report that ANN performs better than linear models when data exhibit high volatility and multicollinearity.


In short, each type of model outperform in different domains. It is apparent that there is no universal model which is suitable for all circumstances. In order to overcome this limitation, various hybrid techniques have been proposed which aim to take advantage of the unique strength of each different type of models.

The common practice in hybrid techniques is to decompose time series data into its linear and nonlinear forms, then use appropriate type of models on them separately. A hybrid ANN-ARIMA model proposed by Zhang \cite{zhang_zhang_2003} achieves more accurate forecast results in time series data as compared to using individual models, in applications such as electricity price forecasting~\cite{adhikari_combination_2013} and water quality prediction~\cite{omer_faruk_hybrid_2010}.  Another successful hybrid ARIMA-ANN technique is presented by Khashei and Bijari \cite{khashei_novel_2011} which defines functional relationship between components. Moreover,  Babu and Reddy~\cite{babu_moving-average_2014} offers a solution to volatility problem in time series data by smoothing out dataset with moving average filter.

Each hybrid method in the literature bring different perspectives to time series forecasting problem. However strong assumptions that these methods make might degenerate their performances if the contrary situations occur.~\cite{zhang_zhang_2003} and~\cite{khashei_novel_2011}  do not decompose time series data into linear and nonlinear components. Rather, they assume that linear component of the data is the output of the ARIMA model. Moreover,~\cite{zhang_zhang_2003} and~\cite{babu_moving-average_2014} assume that the output of their hybrid methods is linear combination of components. However, different datasets can suggest different type of relationships between output and the components.

In this study, we propose  a novel hybrid method for time series forecasting which aims to overcome the limitations of the traditional hybrid methods  by eliminating the need to make strong assumptions. In this method, nature of nonlinearity is first characterized by the help of moving-average (MA) filter, then ARIMA is applied to the  linear component. In the final step, ANN is used to combine the output of ARIMA, the nonlinear component, and the original data.  Three benchmark datasets, the Wolf's sunspot data, the Canadian lynx data, and the British pound/US dollar exchange rate data and an additional public dataset, Turkey Intraday Electricity Market  Price data are used in order to show the effectiveness of the proposed method in time series forecasting.

In addition, we propose an improvement to all hybrid methods mentioned in this paper, including ours, by adding Empirical mode decomposition (EMD) technique~\cite{huang_empirical_1998} to the models. When accuracy results of the hybrid methods with different types of datasets are compared, we observe that accuracy performance gets better with the increasing level of linearity in time series. Then, time series data can be considered as a merge of sub-series which each of them demonstrates more linearity. For this purpose, a well-known multiscale decomposition technique,  EMD is used in the proposed hybrid method. The components achieved by EMD are relatively stationary and have simpler frequency range which make them strongly correlated in themselves. Thus, more accurate predictions can be obtained through the models~\cite{buyuksahin_siu_2018, wang_forecasting_2014}.

The rest of the paper is organized as follows. In the next section, we present existing different time series forecasting methods. In section 3 we present our proposed model. In Section 4, we show the evaluation results of our model and present comparison results with the other methods. In section 5, we present the positive effect of using EMD technique with the hybrid methods and our new method. We show that our method consistently outperforms other linear, nonlinear and even hybrid methods.

 \section{Time series forecasting methods} \label{sec:methods}
 In order to give overall review before we present our method, we want to give a brief information about other time series forecasting methods, such as ARIMA, ANN, and the well-known hybrid methods.

\subsection{Autoregressive Integrated Moving Average Method}
ARIMA is a linear method which means future value of a variable to be forecasted is assumed to be linear function of past observations. As a consequence, time series data that is fed to ARIMA is expected to be linear and stationary. 

The method is composed of three main steps in order:  Identification, estimation parameters and forecasting steps.  In the identification step, stationarity check is performed on given time series data. A stationary time series' statistical characteristics such as mean and autocorrelation patterns constant over time. When a trend or heteroscedasticity is observed on a time serie data, differencing or power transformation methods are generally applied to remove the trend and  make variance stabilize. If stationary property is not satisfied after first attempt, differencing (or power transformation) method  is performed continuously until non-stationary property is disposed. If this operation is performed $d$ times, the integration order of ARIMA model is set to be $d$.  Thereafter, an autoregressive moving average(ARMA) is applied on the resultant data as follows:

Let the actual data value is $y_t$ and random error $\epsilon_t$ at any given time $t$. This actual value $y_t$ is considered as a linear function of the past $p$ observation values, say  $y_{t-1}, y_{t-2}, . . ., y_{t-p}$ and $q$ random erros,  say $\epsilon_t, \epsilon_{t-1}, ..., \epsilon_{t-q}$. The corresponding ARMA equation is given in the following equation: 

\begin{equation}\label{eq:arima} 
    \begin{split}
        y_t = \alpha_1 y_{t-1} + \alpha_2 y_{t-2} + ... + \alpha_p y_{t-p} \\
         + \epsilon_t - \theta_1 \epsilon_{t-1} - \theta_2 \epsilon_{t-2} - ... - \theta_q \epsilon_{t-q}
    \end{split}
\end{equation}

In Equation~\ref{eq:arima},  the coefficients from $\alpha_1$ to $\alpha_p$ are Autoregression coefficients, $\theta_1$ to $\theta_q$ are Moving Average coefficients. Note that random errors $\epsilon_t$  are identically distributed with a mean of zero and a constant variance.

Similar to the $d$ parameter, $p$ and $q$ coefficients are referred to as the orders of the model. When $q$ equals to zero, the model is reduced to AR model of order $p$. If $p$ is equal to zero, the model becomes MA model of order $q$. The main issue in ARIMA modeling is to determine the appropriate model orders $(p,d,q)$. In order to estimate order of the ARIMA model, Box and Jenkins \cite{box_forecasting_2008} proposed to use correlation analyses tools, such as  the autocorrelation function (ACF) and the partial autocorrelation function (PACF). 
When model coefficient estimation is finalized, the future values of the time series data are forecasted using available past data values and estimated model coefficients.

\subsection{Artificial Neural Networks Method}
ANN provides flexible computation framework for nonlinear modeling in wide range of applications. Due to its flexible architecture, number of layers and the  neurons at each layer can be easily varied. In addition, ANN does not require any prior assumption, such as data stationarity, in model building process. Therefore, the network model is largely determined by the characteristics of the data.

The architecture of the most widely used ANN model in time series forecasting, which is also called as multilayer perceptrons, contains three-layers. The neurons of the processing units are acyclically linked. In order to model time series data using such a network, nonlinear function $f$ of $y_t$ sequence from $y_{t-1}$  to $y_{t-N}$ is constructed as shown in the following equation: 
\begin{equation}\label{eq:ann} 
    y_t = w_0 + \sum_{j=1}^{H} w_{j}f (w_{0j} + \sum_{i=1}^{N} w_{ij} y_{t-i}) + e_t
\end{equation}
where, at any given time $t$,  {$w_{ij}$ and $w_j$ are model weights and $H$ and $N$ are the number of hidden and input nodes, respectively. In this equation, $e_t$ corresponds to a noise or error term. The transfer function of the hidden layers $f$ in ANN architecture is generally a sigmoid function.

The power of ANN comes from its flexibility to approximate any continuous function by changing the number of layers $N$ and hidden nodes $H$. 
The choice of number of layers and the nodes at each of them play important role in ANNs' forecasting performance. Large numbers of $N$ and $H$ can give very high training accuracies, but since it tends to memorize the training data, it suffers from overfitting. On the other hand, a too simple network of ANN leads to poor generalization. 
Unfortunately, there is no systematic set of rules to decide the value of these parameters. Thus, extensive number of experiments are required to tune functions and the parameters.

\subsection{Zhang's Hybrid Method}
For time series forecasting, Zhang proposed a hybrid ARIMA-ANN model\cite{zhang_zhang_2003}. According to this model, it is assumed that time series data is a sum of linear and nonlinear components, given in the form of:
\begin{equation}\label{eq:zhang} 
    y_t = L_t + N_t
\end{equation}
where $L_t$ denotes the linear and $N_t$ denotes the nonlinear component.Firstly,  ARIMA is used with the given time series data and linear forecasts are obtained. Residuals from linear component is assumed to contain  only nonlinear relationship. This method uses ARIMA to make forecast from the linear component and ANN from the nonlinear component. Then, these models are combined to improve overall forecasting performance. This method gives better foreceasting accuracy than using ARIMA and ANN methods individually, as seen in the  experimental results in three well-known real data sets - the Wolf’s sunspot data, the Canadian lynx data, and the British pound/US dollar exchange rate data.

\subsection{Khashei and Bijari's Hybrid Method}
For time series forecasting, Khashei and Bijari proposed another hybrid ARIMA-ANN \cite{khashei_novel_2011}. Similar to Zhang's model, this model also assumes that any time series data is composed of linear and nonlinear components. Likewise, ARIMA is used to extract linear component and make forecast on it and residuals, which are  nonlinear components, are fed into ANN along with the original data, and linear forecast of ARIMA output. The difference from the Zhang's model is to avoid the assumption that the relationship between linear and nonlinear components is additive. Rather, this method builds functional relationship between the components  as shown in the following equation:
\begin{equation}\label{eq:khashei} 
    y_t = f (L_t, N_t)
\end{equation}
where $L_t$ is the linear and $N_t$ is the nonlinear component.

In addition, one may not guarantee that the residuals of the linear component may comprise valid nonlinear patterns. Khashei and Bijari suggest that residuals should not put into ANN as an input alone.

\subsection{Babu and Reddy's Hybrid Method}
The hybrid ARIMA-ANN method proposed by Babu and Reddy~\cite{babu_moving-average_2014} integrates moving average filter into hybrid ANN-ARIMA model. Like other methods, this model also assumes that any time series data is composed of linear and nonlinear components. However, this study emphasizes that neither Zhang nor Khashei and Bijari's methods decompose original time series data into its linear and nonlinear components; instead, they use a linear ARIMA model to extract the linear component and the error sequences is assumed to be nonlinear component. On the other hand, this study separates the linear and nonlinear components, then feed them into appropriate methods.

This method tries to fix Moving Average (MA) filter length until \textit{kurtosis} value of the data becomes approximately 3.  Kurtosis is a shape of a probability distribution which measure thickness or heaviness of the tails of a distribution. The kurtosis value is 3 if the data has normal distribution. Shortly, the method aims to find out normal distributed component, which shows low volatility, in time series data by using kurtosis value. When the low-volatile component is separated from the original data, high-volatile component, which is assumed to be nonlinear, is achieved. In the final step, like in Zhang's method, the decomposed components are fed into ARIMA and ANN accordingly and forecast results are summed up to achieve final forecast.

\section{Proposed Method} \label{sec:proposed}
Many decision processes need high forecasting accuracies in time series applications. Although there are numerous available time series models, none of them consistently gives the best results in various situation. There are two main challenges for making an accurate forecast. The first challenge is that underlying data generating process of time series cannot easily identified~\cite{hibon_combine_2005}. The second one is that non-hybrid individual models are generally insufficient to  determine all the characteristics  of the time series~\cite{zhang_zhang_2003}.  Many researches in time series forecasting literature show that hybrid models improve the forecasting performances~\cite{taskaya-temizel_comparative_2005}. By taking the advantage of each individual method in a combined model, error risk of using an inappropriate method is reduced and more accurate results are obtained.

Each hybrid method mentioned in this paper bring different perspectives into time series forecasting problem. However, strong assumptions these methods make might degenerate their performances in certain circumstances. For example, Zhang's and Babu-Reddy's methods assume that the relationship between linear and nonlinear components is additive. If linear and nonlinear components are not additively associated and the relation is different (i.e., it can be multiplicative), the possible complex relationship between components might be overlooked and the forecasting performance might be degenerated. Another assumption that might not always hold is that residuals might not show nonlinear pattern property. Additionally, Zhang's and Khashei-Bijari's methods do not actually decompose data into linear and nonlinear components, but they assume that linear component can be extracted by ARIMA and error sequences shows nonlinear pattern. As a result; such assumptions may lead to low forecasting performances when unexpected scenarios happen. 

\begin{figure}[!ht]
    \centering
    \scalebox{0.7}{
        \includegraphics[center]{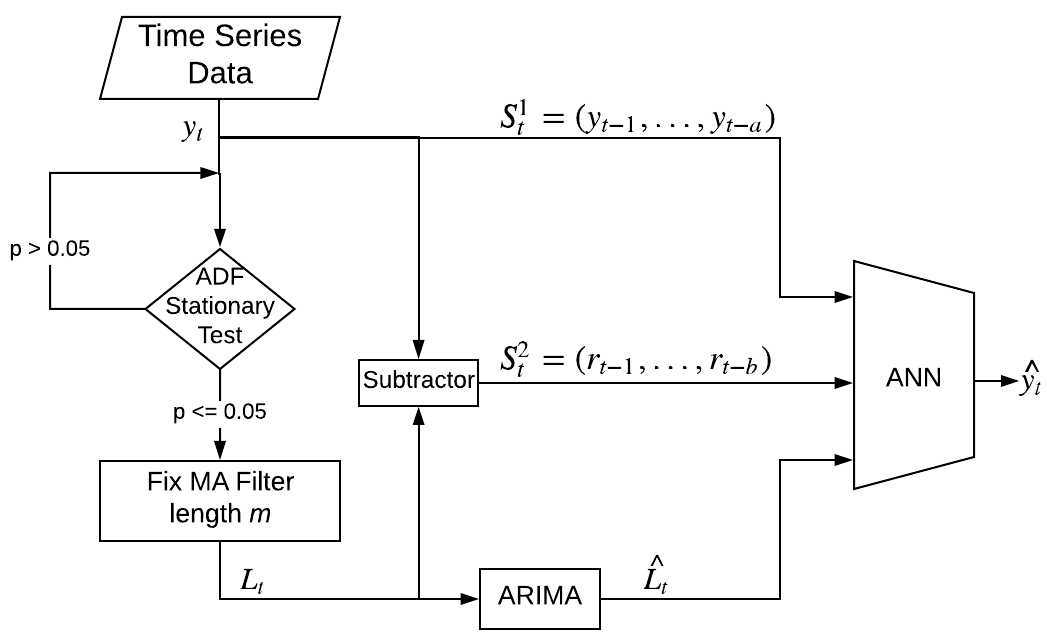}    }
        \caption{Proposed Hybrid Method}
        \label{fig:hybrid_proposed_model}
\end{figure}

In this study, we propose to novel hybrid method for time series forecasting which aims to overcome the limitations of traditional hybrid methods by eliminating strong assumptions. The architecture of the proposed hybrid method is shown in Figure~\ref{fig:hybrid_proposed_model}.

The algorithm starts with data decomposition. In this method time series data $y_t$ is considered as a function of linear $L_t$ and nonlinear $N_t$ components in the same way as given in Equation~\ref{eq:khashei}.


These two components are separated from the original data by using moving average (MA) filter with the length of  $m$,  as given in Equation~\ref{eq:ma_linear}.  While the linear component $l_t$ has low volatility, the residual $r_t$, which is the difference between the original data and the decomposed linear data in Equation~\ref{eq:ma_nonlinear},  shows high fluctuation.
\begin{equation}\label{eq:ma_linear} 
l_t = \frac{1}{m}\sum_{i=t-m+1}^{t}y_i
\end{equation}

\begin{equation}\label{eq:ma_nonlinear} 
r_t = y_t - l_t
\end{equation}

In order for a proper decomposition, the length of the MA filter $m$ has to be adjusted. Augmented Dickey Fuller (ADF) test which  is unit root test can be performed to determine whether a given data series is stationary or not. The existence of a unit root on a given dataset indicates that there is an unpredictable systematic pattern. The more negative ADF test result means the stronger rejection of the existence of unit root for a given time series. Therefore, a negative ADF result implies that the given dataset is stationary. The well accepted threshold is 0.05 which is also used in this study to adjust MA filter length.

After the linear component is achieved with MA filter, a linear  model is constructed as shown in Equation~\ref{eq:model_linear}. The stationary component $l$ is modeled as a linear function of past values of the data series $l_{t-1}, l_{t-2}, ... , l_{t-p}$ and random error series  $\epsilon_{t-1}, \epsilon_{t-2}, ...
, \epsilon_{t-q}$ in Equation~\ref{eq:arima} using ARIMA model.
\begin{equation}\label{eq:model_linear} 
\hat{L_t} =g(l_{t-1}, l_{t-2}, ... , l_{t-p}, \epsilon_{t-1}, \epsilon_{t-2}, ... , \epsilon_{t-q})
\end{equation}
where $g$ is a linear function of ARIMA. 

Finally, nonlinear modeling ANN is used to implement functional relationship between components as indicated in Equation~\ref{eq:khashei}. The past observed data $y_{t-1},y_{t-2}, ... , y_{t-a}$, present ARIMA forecast result of the decomposed stationary data $\hat{L_t}$, and residuals of the data decomposition $r_{t-1},r_{t-2}, ..., r_{t-b}$ are fed to ANN as indicated in Equation~\ref{eq:model_nonlinear}:
\begin{equation}\label{eq:model_nonlinear} 
    \begin{split}
    S^{1}_t &= (y_{t-1},y_{t-2}, ... , y_{t-a}) \\ 
    S^{2}_t &=(r_{t-1},r_{t-2}, ..., r_{t-b}) \\
    \hat{y_t} &=f(S^{1}_t, \hat{L_t}, S^{2}_t ) \\
    \hat{y_t} &=f(y_{t-1},y_{t-2}, …, y_{t-a},\hat{L_t}, r_{t-1},r_{t-2}, ..., r_{t-b})
    \end{split}
\end{equation}
where $f$ is the nonlinear function of ANN, $a$ and $b$ are parameters of the model which show how much we will go back in time to use as features to ANN. Time series data determines how many of those features in the residual path and observed data path are going to be used in the nonlinear model. For example, if the given data does not show volatility, then the residual variable $b$ in Equation~\ref{eq:model_nonlinear} might come out even as zero in tuning process. Likewise, $a$ variable in Equation~\ref{eq:model_nonlinear} is also empirically determined in the tuning process.


The proposed model does not only exploit the unique strength of single models, but also eliminates the three strong assumptions performed by other hybrid methods. Therefore, risk of low forecasting performance in unexpected situations is highly avoided. The competitive performance of our proposed algorithm is shown in our experimental results by using various type of datasets.
\section{Empirical Results}
The performance results of the proposed hybrid method along with the other methods discussed in this paper are evaluated on four different datasets. Three of them are well-known benchmark datasets - the Wolf's sunspot data, the Canadian lynx data, and the British pound/US dollar exchange rate data - which have been widely used in statistics and the
neural network literature~\cite{zhang_zhang_2003, khashei_novel_2011, babu_moving-average_2014, buyuksahin_siu_2018}. The other dataset is publicly available  electricity price of Turkey Intraday Market~\cite{intraday_epias}.  

In the experiments, only one-step-ahead forecasting is considered.  In order to compare accuracy performances, three evaluation metrics are used: Mean Absolute Error (MAE), Mean Squared Error (MSE) and Mean Absolute Scaled Error (MASE) whose formulations are indicated as follows respectively:
\begin{equation}\label{eq:err_metrics}
    \centering
    \begin{split}
        \textrm{MAE} &= \displaystyle\frac{1}{n}\sum_{t=1}^{n}|e_t| \\ 
        \textrm{MSE} &= \displaystyle\frac{1}{n}\sum_{t=1}^{n}e_t^2 \\
        \textrm{MASE} &= \displaystyle\frac{n-1}{n}\frac{\sum_{t=1}^{n}|e_t|}{\sum_{t=2}^{n}|y_t - y_{t-1}|}
    \end{split}
\end{equation}
where $e_t = y_t - \hat{y}_t$ and $y_t$ is the actual data value, $\hat{y}_t$ is the forecasted value at given time $t$.  While MAE specifies the average of the absolute errors over the performed prediction, MSE measures the average of the squared error. Since both MAE and MSE results depend on the scale of the given data, when comparing time series which have different scales, they are not preferable. Therefore, scale-free error metric MASE can be used to compare forecast accuracy between series. 

\begin{table}[ht]
\centering
\caption{Performance Comparison for All Datasets}
\resizebox{\columnwidth}{!}{%
\begin{threeparttable}
\label{tab:results}
\begin{tabular}{clrrrrrr}
\hline
\multicolumn{1}{l}{\textbf{Datasets}} & \textbf{\begin{tabular}[c]{@{}l@{}}Methods\\ \footnotesize Metrics\end{tabular}} & \multicolumn{1}{c}{ANN} & \multicolumn{1}{c}{ARIMA} & \multicolumn{1}{c}{\begin{tabular}[c]{@{}c@{}}Zhang's\\ Method\end{tabular}} & \multicolumn{1}{c}{\begin{tabular}[c]{@{}c@{}}Khashei-Bijari's \\Method\end{tabular}} & \multicolumn{1}{c}{\begin{tabular}[c]{@{}c@{}}Babu-Reddy's\\ Method\end{tabular}} & \multicolumn{1}{c}{\begin{tabular}[c]{@{}c@{}}Proposed\\ Method\end{tabular}} \\ \hline
\multirow{3}{*}{Sunspot}              & \textit{MAE}                                                       & 14.23                   & 13.37                     & 13.14                                                                        & 10.62                                                                        & 11.39                                                                       & \textbf{10.48}                                                                \\ \cline{2-8} 
                                      & \textit{MSE}                                                       & 353.12                  & 306.97                    & 289.31                                                                       & 205.08                                                                       & 239.90                                                                      & \textbf{194.29}                                                               \\ \cline{2-8} 
                                      & \textit{MASE}                                                      & 0.629                   & 0.591                     & 0.581                                                                        & 0.470                                                                        & 0.504                                                                       & \textbf{0.463}                                                                \\ \hline
\multirow{3}{*}{Lynx}                 & \textit{MAE}                                                       & 0.1249                  & 0.1198                    & \textbf{0.1003}                                                              & 0.1025                                                                       & 0.1102                                                                      & 0.1013                                                                        \\ \cline{2-8} 
                                      & \textit{MSE}                                                       & 0.0241                  & 0.0231                    & 0.0173                                                                       & 0.0175                                                                       & 0.0189                                                                      & \textbf{0.0162}                                                               \\ \cline{2-8} 
                                      & \textit{MASE}                                                      & 0.6185                  & 0.5932                    & \textbf{0.4966}                                                              & 0.50757                                                                      & 0.5457                                                                      & 0.5016                                                                        \\ \hline
\multirow{3}{*}{Gbp/Usd}              & \textit{MAE}                                                       & 428.55                  & 435.72                    & 429.52                                                                       & 406.22                                                                       & 436.34                                                                      & \textbf{404.90}                                                               \\ \cline{2-8} 
                                      & \textit{MSE}                                                       & 3.4681                  & 3.5272                    & 3.4496                                                                       & 3.1053                                                                       & 3.5053                                                                      & \textbf{2.9538}                                                               \\ \cline{2-8} 
                                      & \textit{MASE}                                                      & 1.085                   & 1.103                     & 1.087                                                                        & 1.028                                                                        & 1.104                                                                       & \textbf{1.025}                                                                \\ \hline
\multirow{3}{*}{Intraday}             & \textit{MAE}                                                       & 20.10                   & 20.22                     & 19.16                                                                        & 19.79                                                                        & 19.50                                                                       & \textbf{18.81}                                                                \\ \cline{2-8} 
                                      & \textit{MSE}                                                       & 617.46                  & 652.72                    & 594.09                                                                       & 600.93                                                                       & 619.67                                                                      & \textbf{581.38}                                                               \\ \cline{2-8} 
                                      & \textit{MASE}                                                      & 1.048                   & 1.054                     & 0.999                                                                        & 1.031                                                                        & 1.016                                                                       & \textbf{0.980}                                                                \\ \hline

\end{tabular}
\begin{tablenotes}
  \small
  \item \textit{*} MAE and MSE results are multiplied with $10^{-5}$ in Gbp/Usd dataset
\end{tablenotes}
\end{threeparttable}
 }
\end{table}

To assess the forecasting performance of the different methods, each dataset is divided into  training and testing sets. While the training data is used for model development, the test data is used to evaluate the established model. In order to tune the hyper-parameters the methods, the last 20\% of training dataset is used as validation set.

In addition, due to the fact that ANN performs random initialization and produces different results at each run, the methods which include ANN algorithm are executed 50 times and average results are reported. Table~\ref{tab:results} gives the forecasting results of all examined methods on these all four datasets.

\subsection{Forecasts for Sunspot dataset}
The Wolf's sunspot series, which contains annual activity of spots visible on the face of the sun, has been extensively used in numerous linear and nonlinear models~\cite{khashei_novel_2011}. The data includes the annual count of sunspots from 1700 to 1987 (see Figure~\ref{fig:data_sunspot}) giving a total of 288 observations. ADF stationarity test result of the dataset is 0.083 which is greater than the threshold 0.05. This implies that there is a unit root on the dataset, thus the dataset can be regarded as non-stationary time series. 288 observations in the dataset is divided into two samples: 221 observations between 1700-1920 years are considered as training data to develop the model, the last 67 observations between 1921-1987 years are considered as test data and used to evaluate the model performance.
\begin{figure}[!ht]
    \centering
    \scalebox{0.3}{
        \includegraphics[center]{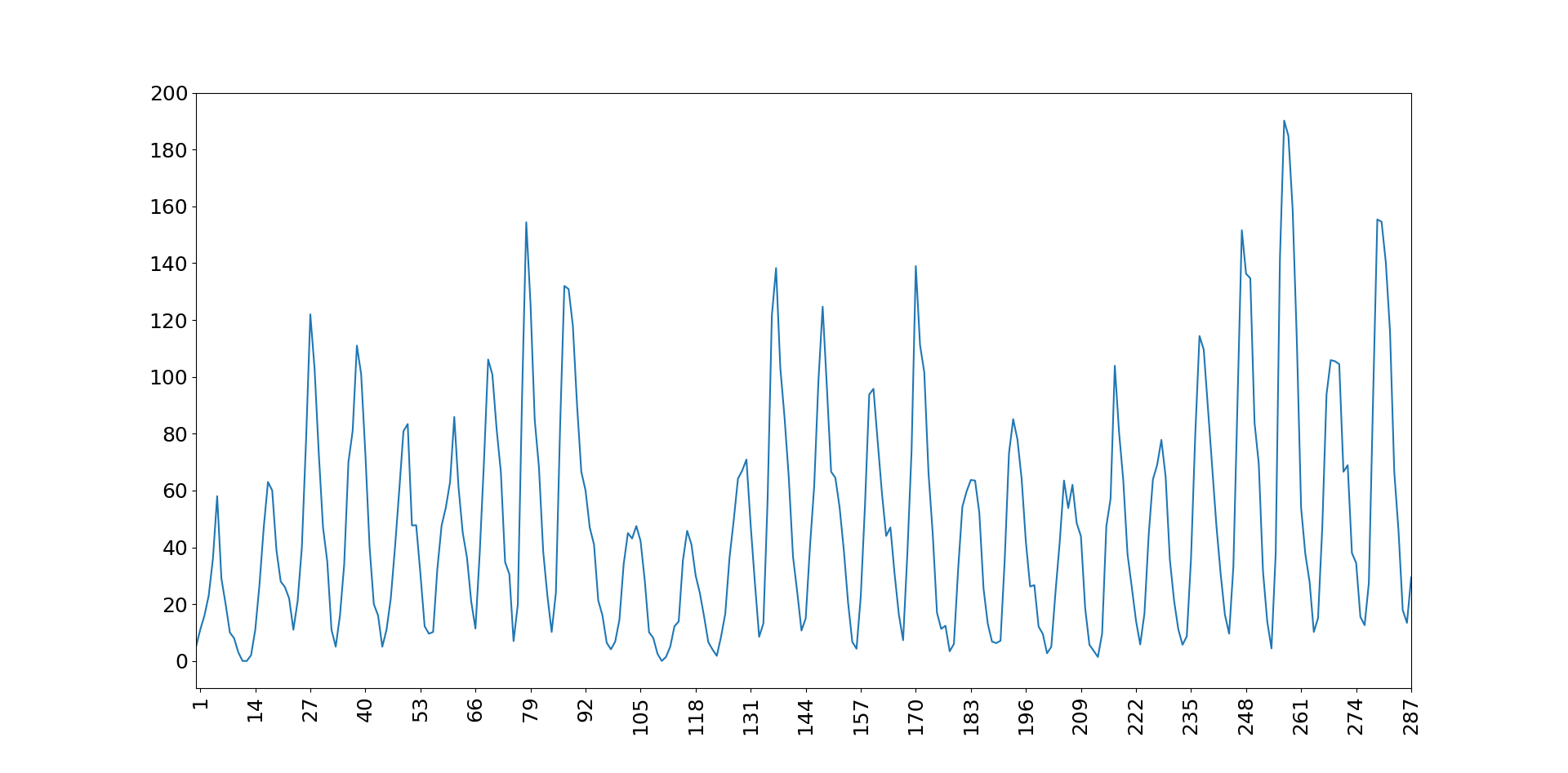}    }
        \caption{Sunspot series (1700-1987)}
        \label{fig:data_sunspot}
\end{figure}

In Sunspot dataset, when ARIMA is individually used as a forecasting method, we also set the order of ARIMA to 9 (AR(9)) as same as the other many studies~\cite{zhang_zhang_2003, khashei_novel_2011, babu_moving-average_2014}.
When ANN is individually used as a forecasting method,  similar to these studies, three layered 4 x 4 x 1 ANN architecture is used which is composed of four input nodes, four hidden nodes, and one output node. 

In the proposed method, the linear component comes out when the MA filter length is 15. After using the filter, the achieved linear component has 0.006 stationary test result which indicates its stationarity, since it is a value less than the threshold, 0.05. 
The best fitted neural network in the final step of the proposed hybrid method has 7 nodes in the input layer where 4 of them are observed values, 2 of them are residuals, and one node is assigned for the result of  linear component forecast. According to our experiments,  when the number of hidden nodes are adjusted to same number of nodes as in the input layer, the best fitted ANN model is achieved.

\begin{figure}[!ht]
    \centerline{
    \scalebox{0.75}{
        \includegraphics{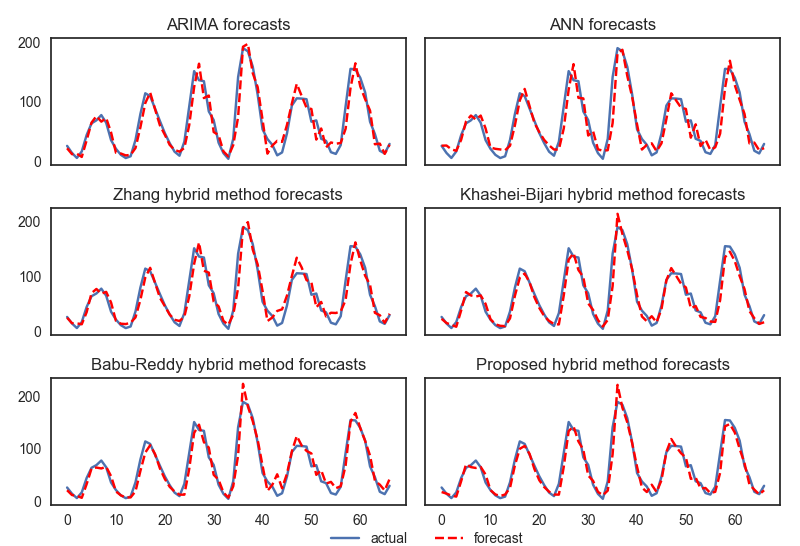}}}
        \caption{Sunspot data forecasts using various methods}
        \label{fig:res_sunspot}
\end{figure}

When numerical results of Sunspot dataset given in Table~\ref{tab:results} are analyzed, individual methods such as ARIMA, ANN have apparently lowest performance as compared to hybrid methods. This suggests that either ARIMA or ANN, when individually used, do not capture all patterns in the data series. Therefore, combining two methods by taking advantage of each of them can be an effective way to overcome this limitation. We indeed observe in Table~\ref{tab:results} that the hybrid methods Zhang, Khashei-Bijari, and  Babu-Reddy methods produce better results as compared to individual ones. However, they produce lower forecasting performance than our proposed 
hybrid method. The assumptions those hybrid methods make can be restricting in many situations as mentioned in Section~\ref{sec:proposed}. Our proposed hybrid method eliminates those assumptions and yields better generalization performance.

\subsection{Forecasts for Lynx dataset}
The lynx dataset, which contains the number of lynx trapped per year in the Mackenzie River district of Northern Canada, is an another extensively analyzed time series data in the literature~\cite{zhang_zhang_2003, khashei_novel_2011}. The data shows a periodicity of approximately 10 years as seen in Figure~\ref{fig:data_lynx}. Moreover, ADF stationary test results of the dataset is 0.056 which implies that the dataset is almost stationary. There are 114 observations in the data, corresponding to the period of 1821-1934. The first 100 observations between 1821-1920 years are considered as training data to develop the model, the last 14 observations between 1921-1934 years are used as test data to evaluate the model performance. In addition, like in other studies~\cite{zhang_zhang_2003, khashei_novel_2011}, the logarithms (to the base 10) of the data are used in the analysis.

\begin{figure}[!ht]
    \centering
    \scalebox{0.3}{
        \includegraphics[center]{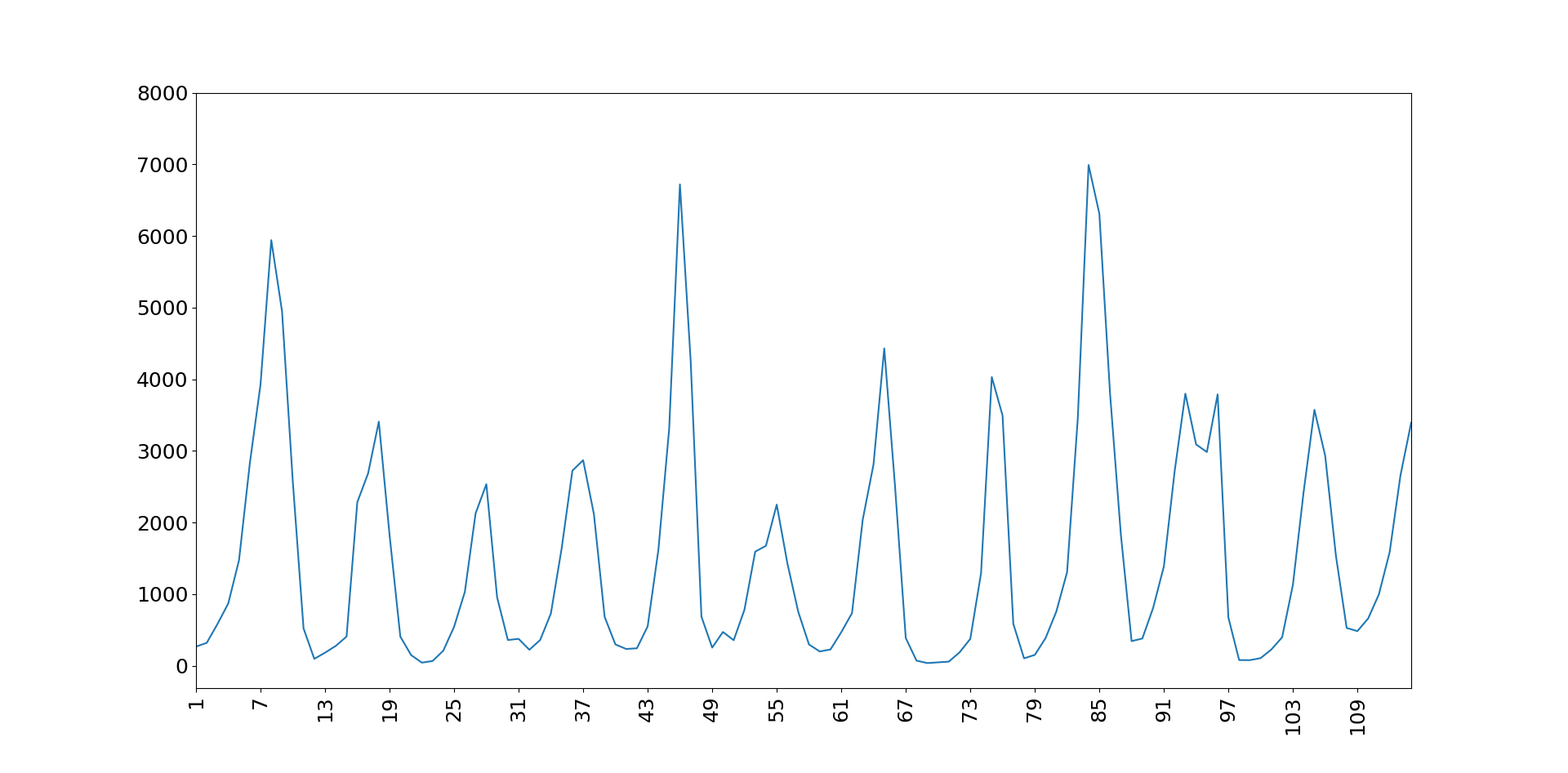}    }
        \caption{Canadian lynx data series (1821-1934)}
        \label{fig:data_lynx}
\end{figure}

When ARIMA  is used as an individual model, we used the AR model of order 12 (AR(12)) for Lynx dataset which is also used by~\cite{zhang_zhang_2003, khashei_novel_2011}. Similar to these studies, three layered 7 X 5 X 1 ANN architecture is used when ANN is individually used as a forecasting method.

In the proposed method, the linear component is extracted from the Lynx dataset when the MA filter length is 5. The relatively short MA filter length was expected, since the ADF test result shows a certain level of  stationarity in the data. As a result of MA filter, the achieved linear component has 0.006 stationary test result which indicates even more stationarity to be properly modeled by ARIMA. The best fitted neural network in the final step of the proposed hybrid method has 9 nodes in the input layer where 5 of them are observed values, 3 of them are residuals, and one node is assigned for the result of linear component forecast.  According to our tuning experiments, when the number of hidden nodes are adjusted to the same number of nodes as in the input layer, the best fitted ANN model is achieved.

\begin{figure}[!ht]
    \centerline{
    \scalebox{0.75}{
        \includegraphics{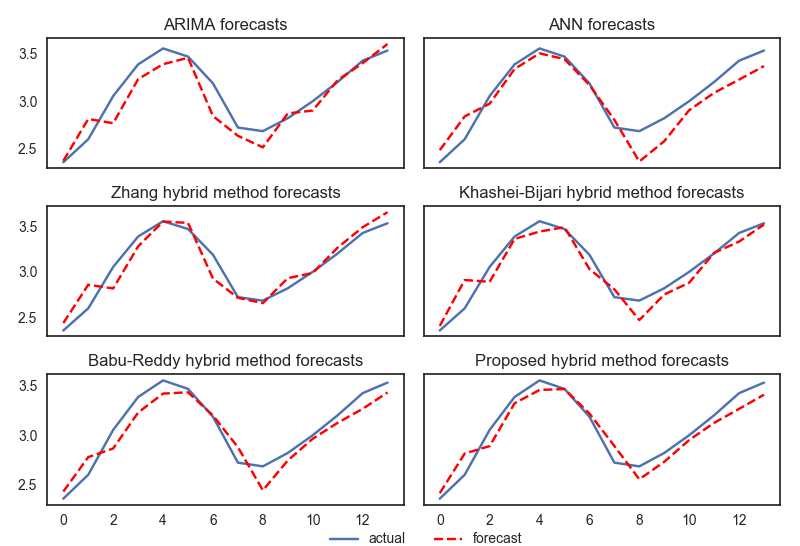}}}
        \caption{Lynx data forecasts using various methods}
        \label{fig:res_lynx}
\end{figure}

In this dataset, among the individually used methods, ARIMA gives better accuracy as compared to ANN in contrast to the Sunspot dataset (see in Table~\ref{tab:results}). This is most likely due to the fact that Lynx dataset is more stationary dataset compared to Sunspot dataset. Due to this relative stationarity, the effect of hybrid methods might not be easily observed. Although we have circumstances which do not necessarily favor the data decomposition and model combination, hybrid methods do not give a lower performance than the individual linear models, and they even provide better results. The highest performance is mostly achieved by our proposed hybrid method. Figure~\ref{fig:res_lynx} compares the actual and forecast values for all examined methods.

\subsection{Forecasts for Gbp/Usd dataset}
The other benchmark dataset is the exchange rate between British pound and US dollar which contains  weekly observation from 1980 to 1993, giving 731 data points in the time series. Predicting exchange rate is an important yet difficult task due to high volatility. ADF stationary test result of the dataset is 0.58 which is highly greater than the threshold 0.05. This implies that the dataset is highly volatile and non-stationary. This non-stationarity can be even seen in the plot, given in Figure~\ref{fig:data_gbpusd}, which shows numerous changing turning points in the series. Similar to other datasets, the experimental setup is same as in previous hybrid studies~\cite{zhang_zhang_2003, khashei_novel_2011}  where data is transformed using natural logarithmic function and separated into two samples. The first 679 observations from 1980-1992 years are considered as training data to develop the model, the last 52 observations between 1992-1993 years are used as test data  to evaluate the model performance.

\begin{figure}[!ht]
    \centering
    \scalebox{0.3}{
        \includegraphics[center]{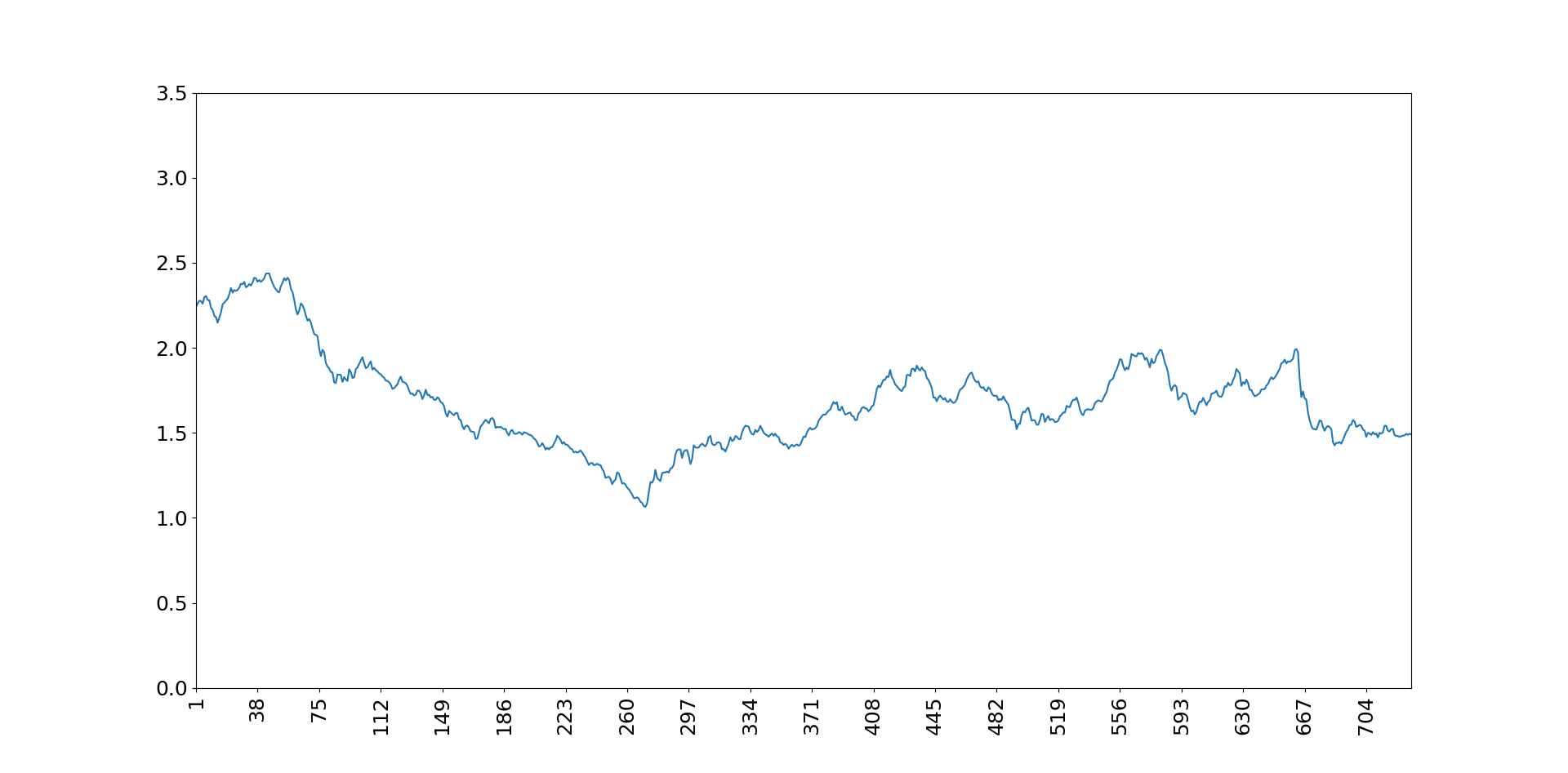}    }
        \caption{Weekly British pound/US dollar exchange rate series (1980-1993)}
        \label{fig:data_gbpusd}
\end{figure}

In this dataset, when ARIMA is individually used,  rather than using regression type of model in the ARIMA itself, random walk model is chosen as best-fitted ARIMA model. This approach has been used by Zhang~\cite{zhang_zhang_2003} and also been suggested by many studies in the exchange rate literature~\cite{moosa_unbeatable_2014}. In this model, the most recent observation is the best guide for the next forecast. When ANN is individually used as a forecasting method, the best fitted ANN is set as three layered 7 x 6 x  1 architecture. 

In the proposed method, in order to decompose this highly volatile data, MA filter length comes out to be 40. As a result of this decomposition, ADF test result of the obtained component is 0.007 which indicates the stationarity of the component. To compute the best final forecast in the proposed model, the ANN is constructed as three layered 9 x 9 x 1 architecture. In this architecture,  input layer is composed of the last 5 of observed values, the last 3 of residuals, and the result of linear component forecast. 

\begin{figure}[!ht]
    \centerline{
    \scalebox{0.75}{
        \includegraphics{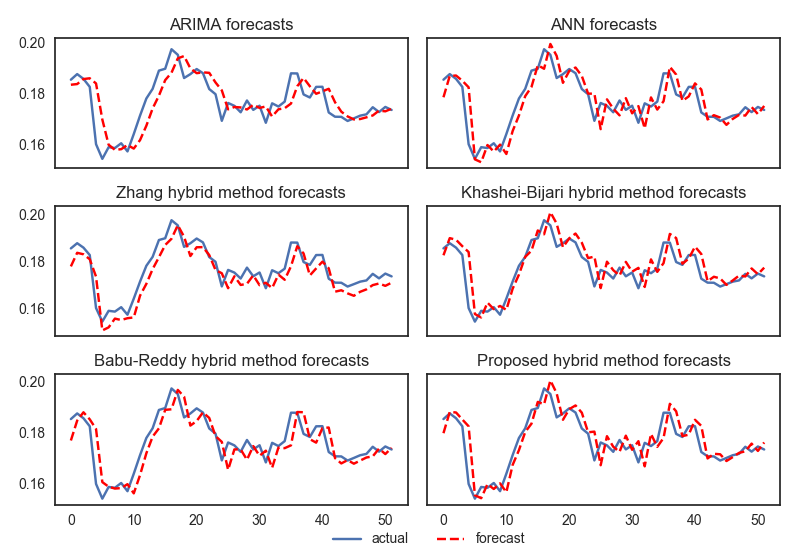}}}
        \caption{Gbp/Usd data forecasts using various methods}
        \label{fig:res_gbpusd}
\end{figure}

Results of the Gbp/Usd dataset forecasts are compared in  Figure~\ref{fig:res_gbpusd}. Both ANN and hybrid methods have much better performance than the individual ARIMA method for a highly fluctuating forecast horizon. The proposed hybrid method is able to capture this volatile pattern much better and outperforms the other methods  in all error metrics.

\subsection{Forecasts for Intraday dataset}
The last analyzed dataset is Turkey intraday electricity market price data which is publicly available~\cite{intraday_epias}. The dataset contains 581 observations which consist of daily averaged prices from July 2015 to December 2017. As compared to  datasets such as Sunspot and Lynx, data pattern in this dataset is also highly fluctuating (see  Figure~\ref{fig:data_intraday}). ADF stationary test result of the dataset shows 0.27 value which is highly greater than the threshold of 0.05. Natural logarithmic transformation is applied on the dataset for scaling purposes.

\begin{figure}[!ht]
    \centering
    \scalebox{0.3}{
        \includegraphics[center]{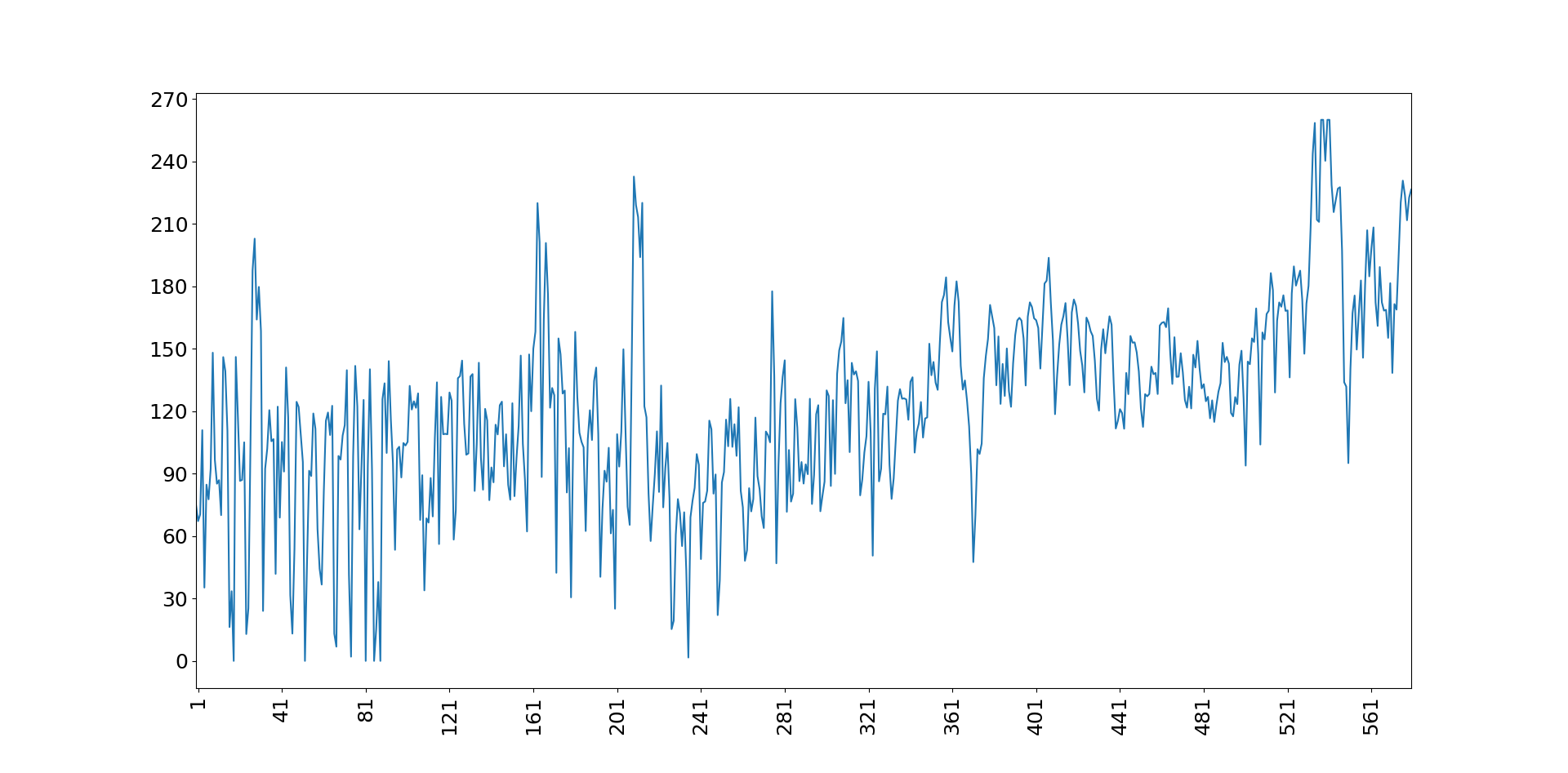}    }
        \caption{Electricity Price of Intraday Market in Turkey (Jul. 2015 - Dec. 2017)}
        \label{fig:data_intraday}
\end{figure}

In this dataset, when ARIMA is individually used as a forecasting method, we found autoregressive model of order 9 (AR(9)) to be the most parsimonious one among all ARIMA models. The best fitted individual ANN model is achieved in three layered 3 X 6 X 1 architecture, after our tuning process.

In this dataset, the length of MA filter which we use in the proposed hybrid method comes out as 6. The obtained linear component after the MA filter has 0.004 ADF stationary test result; this implies that the component can be properly modeled by ARIMA method in our architecture. The best fitted ANN in the final step of the proposed method has 17 x 17 x 1 architecture. Input layer is composed of the last 8 of original data, the last 8 of residuals and the result of linear component forecast.

Similar to Gbp/Usd dataset, Intraday dataset is highly volatile and non-stationary which cannot be effectively modeled by using only a linear model. As can be seen in Table~\ref{tab:results}, ANN and hybrid methods significantly outperform the individual linear model of ARIMA. Furthermore, the proposed method gives remarkably superior accuracy as compared to other hybrid methods in all error metrics. The comparison of the actual and forecast values for all examined method are given in Figure~\ref{fig:res_intraday}.

\begin{figure}[!ht]
    \centerline{
    \scalebox{0.75}{
        \includegraphics{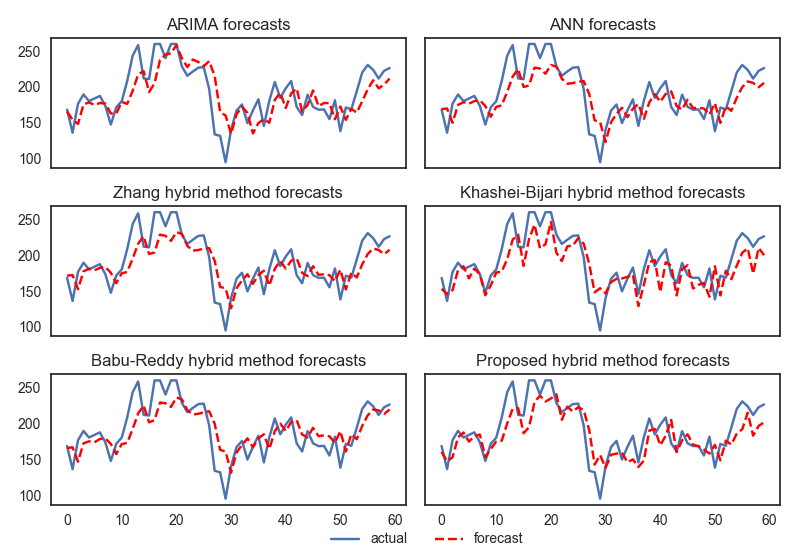}}}
        \caption{Intraday data forecasts using various methods}
        \label{fig:res_intraday}
\end{figure}

\section{Discussion and Improvement}
There are several important results obtained in our experiments Firstly, when  individual methods' results are compared among themselves, we see that ARIMA outperforms ANN for the datasets which present more linearity and vice versa (see Table~\ref{tab:results}). 
Moreover, hybrid methods have better performance as compared to individual ones especially in more fluctuating datasets. Finally, the assumptions made by other hybrid methods degenerate the forecasting performance when unexpected situations occur in the data. Our proposed hybrid method which avoids these assumptions apparently creates more general models and outperforms the other examined methods.

\begin{figure}[!ht]
    \centering
    \scalebox{0.30}{
        \includegraphics[center]{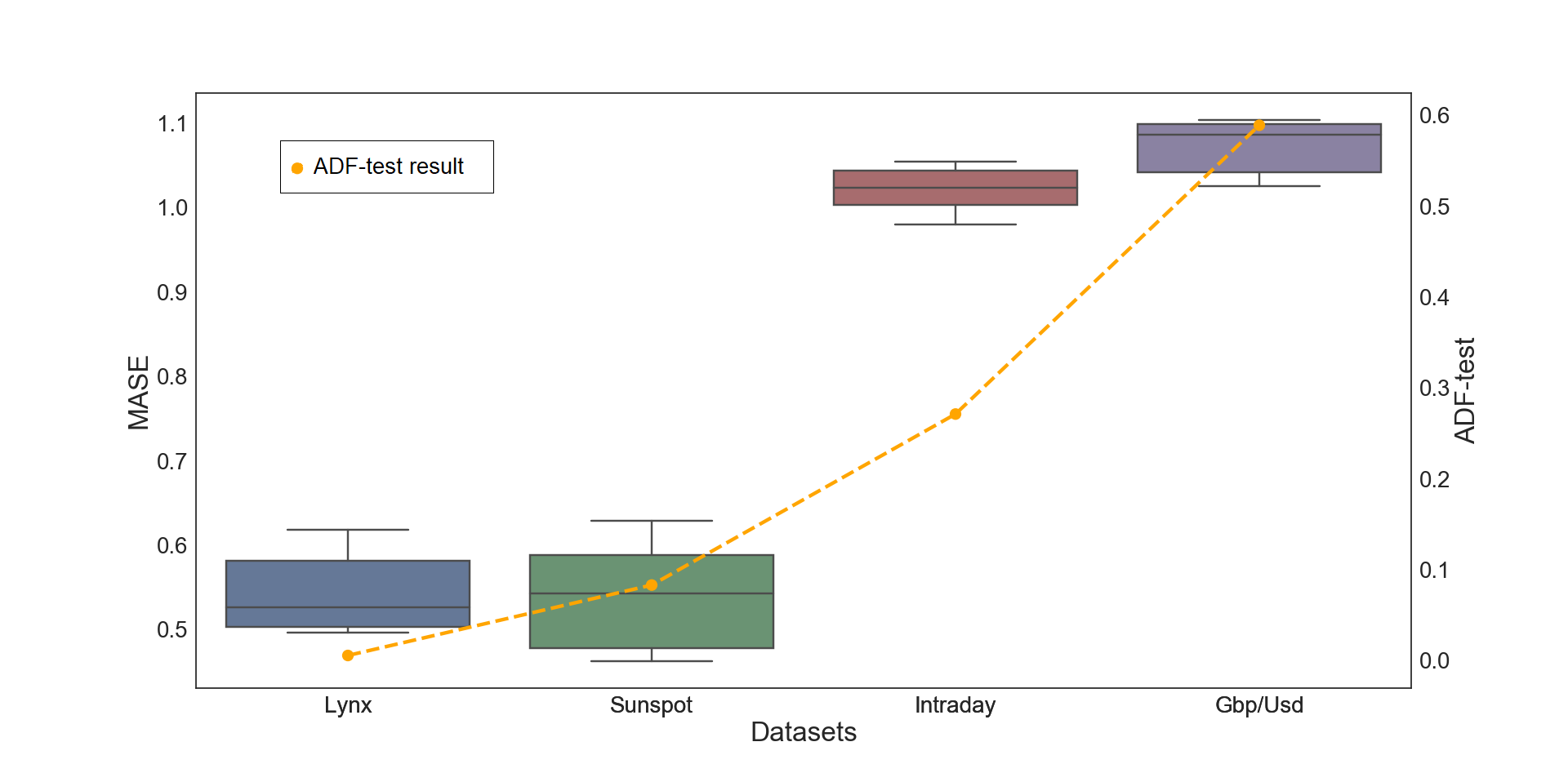}}
        \caption{Comparison of MASE distribution with ADF test results for all datasets}
        \label{fig:mase_vs_adf}
\end{figure}
Figure~\ref{fig:mase_vs_adf} compares the  distribution of MASE results of datasets with ADF test results of the corresponding dataset. In this figure, boxplots are drawn by using MASE values of all examined methods for each dataset, presented in Table~\ref{tab:results}. When we compare error results among time series data by using a scale-invariant error metric MASE, it is observed that the more non-stationarity in a dataset leads to a higher error value. For example, Lynx dataset, which turns out to be the most linear among all datasets according to ADF test results, has the lowest MASE results. On the other hand, Gbp/Usd dataset, which shows the most non-linearity according to ADF test results, has the highest MASE results. As a result of this, we can conclude that having more regular data distribution in a time series leads to more accurate results in forecasting. This conclusion motivates us to propose an improvement on our already best performing proposed hybrid method. This improved method aims to produce more stationary subseries from given time series by using a multi-scale decomposition technique. Then, those achieved linear subseries can be modeled with a higher accuracy using the proposed hybrid method.

In the literature there are several multi-scale decomposition methods such as Empirical Mode Decomposition (EMD),  Wavelet Packet Decomposition (WPD), Fourier Transform (FT) and etc.~\cite{huang_applications_2003}.  Since EMD does not make a priori assumption about the given time series and preserves time scale of the data throughout the decomposition, it is a more prefer technique than the others for decomposing  time series~\cite{huang_empirical_1998}. 

\begin{figure}[!ht]
    \centering
    \scalebox{0.6}{
        \includegraphics[center]{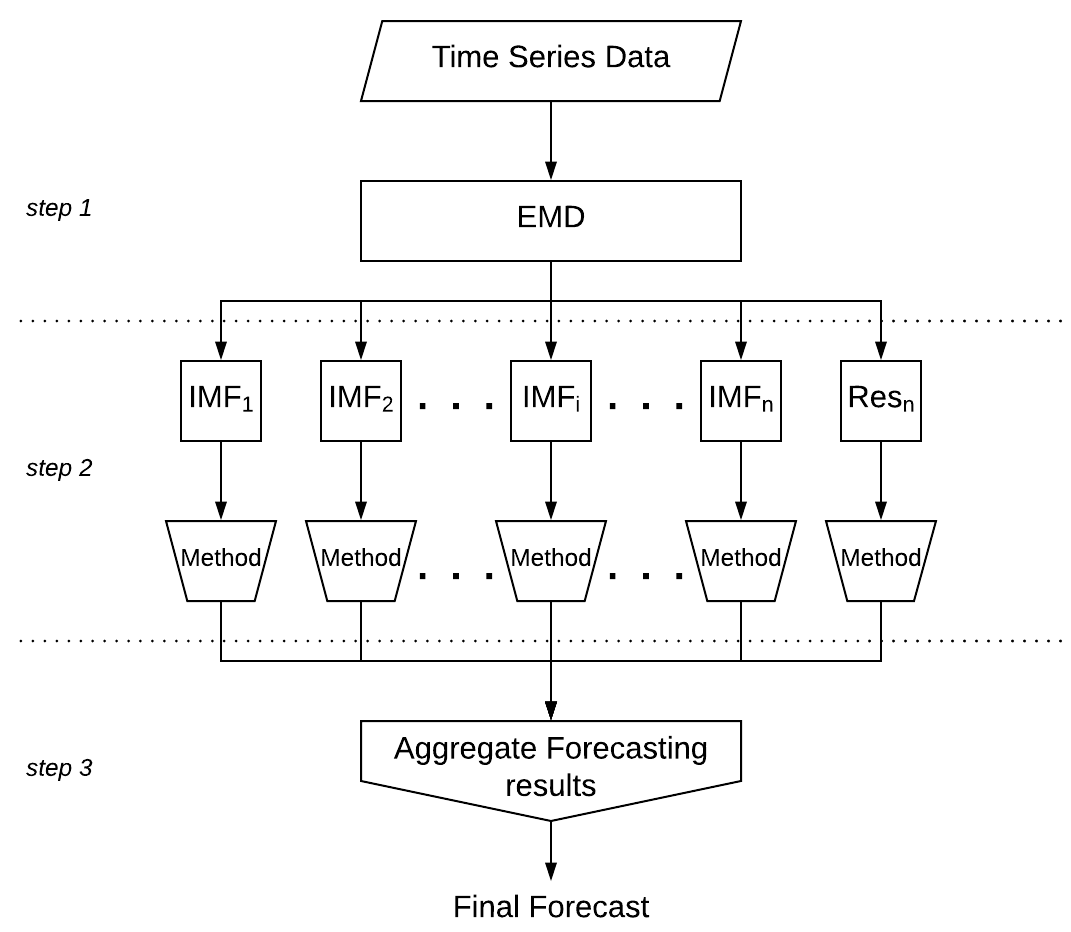}}
        \caption{The hybrid architecture using EMD}
        \label{fig:arch_emd}
\end{figure}

The main principle of EMD is to decompose a given time series data into a sum of several subseries. Those subseries are  called \textit{Intrinsic Mode Functions (IMFs)} and the remaining component after subtracting the summation of IMFs from the original data is called \textit{residue}. These subseries have two important properties which allow them to be easily modeled: Each subseries has its own local characteristic time scale and they are relatively stationary subseries.  Let $y(t)$ be a given time series data, and then the EMD calculation can be described as follows:
\begin{equation}\label{eq:emd} 
	y(t) = \sum_{i=1}^{n} IMF_i(t) + R_n(t)
\end{equation}
where $IMF_i(t)$  $(i = 1, 2, . . ., n)$ represents the different subseries, and $R_n(t)$ is the residue after summation of $n$ IMFs 
are subtracted from the original data.

The EMD-based methods includes three main steps, as seen in Figure~\ref{fig:arch_emd}: In the first step, the original time series data is decomposed into IMFs. In the second step, forecasting is performed by usinf our hybr'dd method on each IMF. In the last step, forecast results of each individual model are summed up to achieve the final forecast of the original time series. We use an additive function in the end to capture the additive relation of IMFs with the original data.


\begin{table}[h]
\centering
\caption{Performance Comparison of methods using EMD for All Datasets}
\resizebox{\columnwidth}{!}{%
\begin{threeparttable}
\label{tab:emdresults}
\begin{tabular}{clrrrrrr}
\hline
\multicolumn{1}{l}{\textbf{Datasets}} & \textbf{\begin{tabular}[c]{@{}l@{}}Methods\\ \footnotesize Metrics\end{tabular}} & \multicolumn{1}{c}{ANN} & \multicolumn{1}{c}{ARIMA} & \multicolumn{1}{c}{\begin{tabular}[c]{@{}c@{}}Zhang's\\ Method\end{tabular}} & \multicolumn{1}{c}{\begin{tabular}[c]{@{}c@{}}Khashei-Bijari's\\Method\end{tabular}} & \multicolumn{1}{c}{\begin{tabular}[c]{@{}c@{}}Babu-Reddy's\\Method\end{tabular}} & \multicolumn{1}{c}{\begin{tabular}[c]{@{}c@{}}Proposed\\ Method\end{tabular}} \\ \hline
\multirow{3}{*}{Sunspot}              & \textit{MAE}                                                       & 8.33                    & 7.72                      & 7.46                                                                         & 7.76                                                                         & 7.92                                                                        & \textbf{7.28}                                                                 \\ \cline{2-8} 
                                      & \textit{MSE}                                                       & 120.14                  & 99.07                     & 90.04                                                                        & 99.17                                                                        & 100.64                                                                      & \textbf{87.86}                                                                \\ \cline{2-8} 
                                      & \textit{MASE}                                                      & 0.368                   & 0.341                     & 0.330                                                                        & 0.343                                                                        & 0.350                                                                       & \textbf{0.322}                                                                \\ \hline
\multirow{3}{*}{Lynx}                 & \textit{MAE}                                                       & 0.0912                  & 0.0772                    & 0.0764                                                                       & 0.0782                                                                       & 0.0788                                                                      & \textbf{0.0760}                                                               \\ \cline{2-8} 
                                      & \textit{MSE}                                                       & 0.01318                 & 0.00996                   & 0.00995                                                                      & 0.00998                                                                      & 0.01014                                                                     & \textbf{0.00923}                                                              \\ \cline{2-8} 
                                      & \textit{MASE}                                                      & 0.4516                  & 0.3822                    & 0.3783                                                                       & 0.3872                                                                       & 0.3902                                                                      & \textbf{0.3763}                                                               \\ \hline
\multirow{3}{*}{Gbp/Usd}              & \textit{MAE}                                                       & 190.08                  & 146.03                    & 142.92                                                                       & 146.11                                                                       & 147.18                                                                      & \textbf{141.38}                                                               \\ \cline{2-8} 
                                      & \textit{MSE}                                                       & 0.5610                  & 0.3578                    & 0.3479                                                                       & 0.3581                                                                       & 0.3633                                                                      & \textbf{0.3285}                                                               \\ \cline{2-8} 
                                      & \textit{MASE}                                                      & 0.4812                  & 0.3697                    & 0.3618                                                                       & 0.3699                                                                       & 0.3726                                                                      & \textbf{0.3579}                                                               \\ \hline
\multirow{3}{*}{Intraday}             & \textit{MAE}                                                       & 10.96                   & 10.79                     & 10.37                                                                        & 9.88                                                                         & 10.05                                                                       & \textbf{8.93}                                                                 \\ \cline{2-8} 
                                      & \textit{MSE}                                                       & 182.86                  & 158.64                    & 158.88                                                                       & 162.55                                                                       & 165.41                                                                      & \textbf{130.79}                                                               \\ \cline{2-8} 
                                      & \textit{MASE}                                                      & 0.571                   & 0.562                     & 0.540                                                                        & 0.515                                                                        & 0.524                                                                       & \textbf{0.456}                                                                \\ \hline
\end{tabular}
\begin{tablenotes}
  \small
  \item \textit{*} MAE and MSE results are multiplied with $10^{-5}$ in Gbp/Usd dataset
\end{tablenotes}
\end{threeparttable}
 }
\end{table}

The EMD-based methods are evaluated on the same datasets by using the 
experimental setup in Figure~\ref{fig:arch_emd}. In order to evaluate the effect of EMD, all examined methods are executed in the second step of the algorithm (see Figure~\ref{fig:arch_emd}). Table~\ref{tab:emdresults} gives the forecasting results of all examined methods with EMD on four all datasets. When these results are compared with the previous ones showed in Table~\ref{tab:results}, the methods with EMD give significantly higher accuracies. The percentage improvement for all datasets are presented in Table~\ref{tab:imptables}. The improvements at each dataset varies between 23\% and 89\% for all error metrics. We also provide a bar chart (see Figure~\ref{fig:mase_vs_emd}) that shows the MASE results for each method which are averaged over the results of all datasets. The chart indicates that methods with EMD achieve remarkably less error in their forecasts. Further analysis shows that EMD-based methods give even greater improvements in the accuracies for non-stationary datasets. This is most likely due to the fact that EMD is able to resolve high volatility problem in time series data. For example, while our hybrid method with EMD improves MASE results of relatively stationary datasets Sunspot and Lynx 30\% and 28\%  respectively, this improvement jump over 50\% in Gbp/Usd and Intraday datasets which are highly non-stationary. Another indication of EMD's capability of solving volatility problem is that ARIMA is able to achieve better results than ANN for all datasets (see Table~\ref{tab:imptables}), although ANN without using EMD  was better in non-stationary datasets (see Table~\ref{tab:results}).

\begin{figure}[!ht]
    \centering
    \scalebox{0.3}{
        \includegraphics[center]{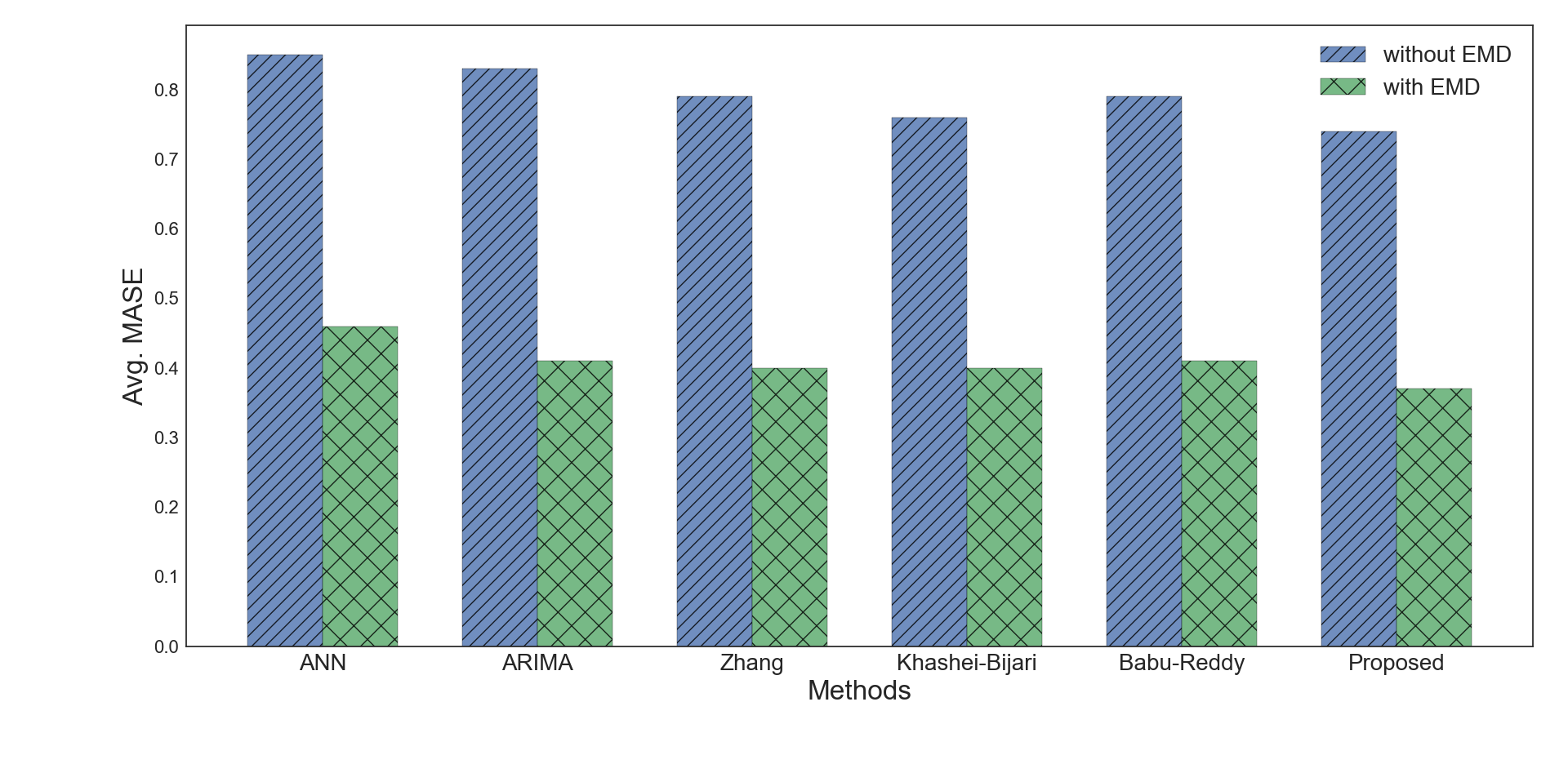}}
        \caption{Average MASE results of the methods with/without using EMD}
        \label{fig:mase_vs_emd}
\end{figure}

In final, we want to point out that our proposed hybrid method with EMD gives the best results as compared to other methods (see Table~\ref{tab:imptables}). Subseries obtained from EMD are relatively stationary as compared to original data however, they still show fluctuations in their frequency range. Performing one more decomposition on these subseries using MA filter and constituting functional relation between the stationary part, residuals, and the original data values outperforms all other remaining methods.

\begin{table}[h]
\centering
\caption{Percentage improvement in the methods when EMD is used}
\resizebox{\columnwidth}{!}{%
\label{tab:imptables}
\begin{tabular}{clcccccc}
\hline
\multicolumn{1}{l}{\textbf{Datasets}} & \textbf{\begin{tabular}[c]{@{}l@{}}Methods\\ \\ \footnotesize{Metrics} \end{tabular}} & \begin{tabular}[c]{@{}c@{}}ANN\\ \\ (\%)\end{tabular} & \begin{tabular}[c]{@{}c@{}}ARIMA\\ \\ (\%)\end{tabular} & \begin{tabular}[c]{@{}c@{}}Zhang's\\ Method\\ (\%)\end{tabular} & \begin{tabular}[c]{@{}c@{}}Khashei-Bijari's\\ Method\\ (\%)\end{tabular} & \begin{tabular}[c]{@{}c@{}}Babu-Reddy's\\ Method\\ (\%)\end{tabular} & \begin{tabular}[c]{@{}c@{}}Proposed\\ Method\\ (\%)\end{tabular} \\ \hline
\multirow{3}{*}{Sunspot}              & \textit{MAE}                                                       & 41.1                                                  & 42.2                                                    & 43.2                                                            & 26.9                                                                     & 30.4                                                                 & 30.5                                                             \\ \cline{2-8} 
                                      & \textit{MSE}                                                       & 65.9                                                  & 67.7                                                    & 68.8                                                            & 51.6                                                                     & 58.0                                                                 & 54.7                                                             \\ \cline{2-8} 
                                      & \textit{MASE}                                                      & 41.1                                                  & 42.2                                                    & 43.2                                                            & 26.9                                                                     & 30.4                                                                 & 30.5                                                             \\ \hline
\multirow{3}{*}{Lynx}                 & \textit{MAE}                                                       & 26.9                                                  & 35.5                                                    & 25.1                                                            & 23.7                                                                     & 28.4                                                                 & 28.4                                                             \\ \cline{2-8} 
                                      & \textit{MSE}                                                       & 45.6                                                  & 57.1                                                    & 42.7                                                            & 43.4                                                                     & 46.5                                                                 & 43.2                                                             \\ \cline{2-8} 
                                      & \textit{MASE}                                                      & 26.9                                                  & 35.5                                                    & 25.1                                                            & 23.7                                                                     & 28.4                                                                 & 28.4                                                             \\ \hline
\multirow{3}{*}{Gbp/Usd}              & \textit{MAE}                                                       & 55.6                                                  & 66.4                                                    & 62.3                                                            & 64.0                                                                     & 66.2                                                                 & 65.0                                                             \\ \cline{2-8} 
                                      & \textit{MSE}                                                       & 83.8                                                  & 90.0                                                    & 90.1                                                            & 88.7                                                                     & 89.7                                                                 & 89.1                                                             \\ \cline{2-8} 
                                      & \textit{MASE}                                                      & 55.6                                                  & 66.4                                                    & 62.3                                                            & 64.0                                                                     & 66.2                                                                 & 65.0                                                             \\ \hline
\multirow{3}{*}{Intraday}             & \textit{MAE}                                                       & 45.4                                                  & 46.6                                                    & 44.4                                                            & 50.0                                                                     & 48.4                                                                 & 52.5                                                             \\ \cline{2-8} 
                                      & \textit{MSE}                                                       & 70.3                                                  & 75.6                                                    & 73.2                                                            & 72.9                                                                     & 73.3                                                                 & 77.5                                                             \\ \cline{2-8} 
                                      & \textit{MASE}                                                      & 45.4                                                  & 46.6                                                    & 44.4                                                            & 50.0                                                                     & 48.4                                                                 & 52.5                                                             \\ \hline
\end{tabular}
 }
\end{table}
\section{Conclusions}
Time series forecasting is an important yet often a challenging task used in many different application domains. The studies in the litreture mainly focus on either linear or nonlinear modelings individually or a combination of them. While linear models such as ARIMA gives better forecasting accuracy with stationary time series data, nonlinear methods such as ANN is more appropriate for non-stationary datasets. 
In order to take advantage of the unique strength of each different type of methods in a more general setting, hybrid methods are proposed.  Hybrid methods basically use linear and nonlinear modeling, ARIMA and ANN respectively on the corresponding decomposed components and then combine the results. Hybrid ARIMA-ANN methods give better results in general as compared to the cases where they are individually used. However, they generally suffer from the assumptions they make while constructing their model. These assumptions lead to produce inconsistent results and give low accuracies in overall if unexpected situations occur.
 
In this study, a new hybrid ARIMA-ANN model based forecasting method is proposed to overcome three main assumptions made by traditional hybrid ARIMA-ANN models. Firstly, the proposed method removes the assumption that the linear component is the ARIMA model output of the given data. Rather, it extracts the linear component by using MA filter. It is known that data showing linear characteristics can be more accurately modeled by linear methods. Therefore, properly decomposed data yields more accurate linear forecasting and consequently more accurate final results in the hybrid methods. Secondly, the proposed method does not directly model residuals via a nonlinear method ANN, since the assumption of that residuals might comprise valid nonlinear patterns, does not always hold. Thirdly, the proposed method  does not restrict linear and nonlinear component modeling and also combining the results of them. Rather, it can capture structures of the linear and nonlinear components in a better way, and produces more general models than those existing hybrid models.


In the light of our experimental results, we can conclude that forecasting performance gets better if more stationary time series data is provided. This result motives us to make original time series data more stationary in order to improve accuracy results. We show that when EMD multi-scale data decomposition is combined with all examined methods, accuracy results can be remarkably improved. Our experimental results indicate that our hybrid method with EMD gives remarkably superior accuracy as compared to all other examined methods.



\bibliography{elsarticle-template}

\begin{thebibliography}{10}
\expandafter\ifx\csname url\endcsname\relax
  \def\url#1{\texttt{#1}}\fi
\expandafter\ifx\csname urlprefix\endcsname\relax\def\urlprefix{URL }\fi
\expandafter\ifx\csname href\endcsname\relax
  \def\href#1#2{#2} \def\path#1{#1}\fi

\bibitem{ph.d_sas_2018}
J.~C. Brocklebank, D.~A. Dickey, B.~Choi, {SAS} for {Forecasting} {Time}
  {Series}, {3rd} {Edition}, SAS Institute, 2018.

\bibitem{contreras_arima_2002}
J.~Contreras, R.~Espinola, F.~Nogales, A.~J.~Conejo, {ARIMA} {Models} to
  {Predict} {Next}-{Day} {Electricity} {Prices}, Power Engineering Review, IEEE
  22 (2002) 57--57.
\newblock \href {http://dx.doi.org/10.1109/MPER.2002.4312577}
  {\path{doi:10.1109/MPER.2002.4312577}}.

\bibitem{gonzalez-romera_monthly_2006-1}
E.~Gonzalez-Romera, M.~A. Jaramillo-Moran, D.~Carmona-Fernandez, Monthly
  {Electric} {Energy} {Demand} {Forecasting} {Based} on {Trend} {Extraction},
  IEEE Transactions on Power Systems 21~(4) (2006) 1946--1953.
\newblock \href {http://dx.doi.org/10.1109/TPWRS.2006.883666}
  {\path{doi:10.1109/TPWRS.2006.883666}}.

\bibitem{k_p_modelling_2016}
V.~K~P, P.~Sahu, B.~Dhekale, P.~Mishra, Modelling and {Forecasting} {Sugarcane}
  and {Sugar} {Production} in {India}, Indian Journal of Economics and
  Development 12 (2016) 71.
\newblock \href {http://dx.doi.org/10.5958/2322-0430.2016.00009.3}
  {\path{doi:10.5958/2322-0430.2016.00009.3}}.

\bibitem{wang_improving_2015}
W.-c. Wang, K.-w. Chau, D.-M. Xu, X.-Y. Chen, Improving {Forecasting}
  {Accuracy} of {Annual} {Runoff} {Time} {Series} {Using} {ARIMA} {Based} on
  {EEMD} {Decomposition}, Water Resources Management 29 (2015) 2655--2675.
\newblock \href {http://dx.doi.org/10.1007/s11269-015-0962-6}
  {\path{doi:10.1007/s11269-015-0962-6}}.

\bibitem{box_forecasting_2008}
G.~E.~P. Box, G.~M. Jenkins, G.~C. Reinsel, G.~E.~P. Box, G.~M. Jenkins, G.~C.
  Reinsel, Forecasting, in: Time {Series} {Analysis}, John Wiley \& Sons, Inc.,
  2008, pp. 137--191.
\newblock \href {http://dx.doi.org/10.1002/9781118619193.ch5}
  {\path{doi:10.1002/9781118619193.ch5}}.

\bibitem{ediger_arima_2007}
V.~Ediger, S.~Akar, {ARIMA} forecasting of primary energy demand by fuel in
  {Turkey}, Energy Policy 35 (2007) 1701--1708.
\newblock \href {http://dx.doi.org/10.1016/j.enpol.2006.05.009}
  {\path{doi:10.1016/j.enpol.2006.05.009}}.

\bibitem{lapedes_nonlinear_1987}
A.~Lapedes, R.~Farber,
  \href{https://www.osti.gov/scitech/biblio/5470451}{Nonlinear {Signal}
  {Processing} {Using} {Neural} {Networks}: {Prediction} and {System}
  {Modelling}}, 1. IEEE international conference on neural networks, 1987, San
  Diego.
\newline\urlprefix\url{https://www.osti.gov/scitech/biblio/5470451}

\bibitem{chen_using_2009}
W.-S. Chen, Y.-K. Du, Using {Neural} {Networks} and {Data} {Mining}
  {Techniques} for the {Financial} {Distress} {Prediction} {Model}, Expert
  Syst. Appl. 36~(2) (2009) 4075--4086.
\newblock \href {http://dx.doi.org/10.1016/j.eswa.2008.03.020}
  {\path{doi:10.1016/j.eswa.2008.03.020}}.

\bibitem{singhal_electricity_2011}
D.~Singhal, K.~S. Swarup, Electricity price forecasting using artificial neural
  networks, International Journal of Electrical Power \& Energy Systems 33~(3)
  (2011) 550--555.
\newblock \href {http://dx.doi.org/10.1016/j.ijepes.2010.12.009}
  {\path{doi:10.1016/j.ijepes.2010.12.009}}.

\bibitem{ardabili_computational_2018}
S.~F. Ardabili, B.~Najafi, S.~Shamshirband, B.~M. Bidgoli, R.~C. Deo, K.~wing
  Chau, Computational intelligence approach for modeling hydrogen production: a
  review, Engineering Applications of Computational Fluid Mechanics 12~(1)
  (2018) 438--458.
\newblock \href {http://dx.doi.org/10.1080/19942060.2018.1452296}
  {\path{doi:10.1080/19942060.2018.1452296}}.

\bibitem{chang_novel_2009}
B.~R. Chang, H.~F. Tsai, Novel hybrid approach to data-packet-flow prediction
  for improving network traffic analysis, Applied Soft Computing 9~(3) (2009)
  1177--1183.
\newblock \href {http://dx.doi.org/10.1016/j.asoc.2009.03.003}
  {\path{doi:10.1016/j.asoc.2009.03.003}}.

\bibitem{cybenko_approximation_1992}
G.~Cybenko, Approximation by superpositions of a sigmoidal function,
  Mathematics of Control, Signals and Systems 5~(4) (1992) 455--455.
\newblock \href {http://dx.doi.org/10.1007/BF02134016}
  {\path{doi:10.1007/BF02134016}}.

\bibitem{hornik_multilayer_1989}
K.~Hornik, M.~Stinchcombe, H.~White, Multilayer feedforward networks are
  universal approximators, Neural Networks 2~(5) (1989) 359--366.
\newblock \href {http://dx.doi.org/10.1016/0893-6080(89)90020-8}
  {\path{doi:10.1016/0893-6080(89)90020-8}}.

\bibitem{zhang_zhang_2003}
P.~Zhang, {Time} {Series} {Forecasting} {Using} a {Hybrid} {ARIMA} and {Neural}
  {Network} {Model}. {Neurocomputing} 50, 159-175, Neurocomputing 50 (2003)
  159--175.
\newblock \href {http://dx.doi.org/10.1016/S0925-2312(01)00702-0}
  {\path{doi:10.1016/S0925-2312(01)00702-0}}.

\bibitem{foster_neural_1992}
W.~R. Foster, F.~Collopy, L.~H. Ungar, Neural network forecasting of short,
  noisy time series, Computers \& Chemical Engineering 16~(4) (1992) 293--297.
\newblock \href {http://dx.doi.org/10.1016/0098-1354(92)80049-F}
  {\path{doi:10.1016/0098-1354(92)80049-F}}.

\bibitem{brace_comparison_1991}
M.~C. Brace, J.~Schmidt, M.~Hadlin, Comparison of the forecasting accuracy of
  neural networks with other established techniques, in: Proceedings of the
  {First} {International} {Forum} on {Applications} of {Neural} {Networks} to
  {Power} {Systems}, 1991, pp. 31--35.
\newblock \href {http://dx.doi.org/10.1109/ANN.1991.213493}
  {\path{doi:10.1109/ANN.1991.213493}}.

\bibitem{aras_new_2016}
S.~Aras, Ä.~D. Kocakoç, A new model selection strategy in time series
  forecasting with artificial neural networks: {IHTS}, Neurocomputing 174
  (2016) 974--987.
\newblock \href {http://dx.doi.org/10.1016/j.neucom.2015.10.036}
  {\path{doi:10.1016/j.neucom.2015.10.036}}.

\bibitem{denton_how_good_1995}
{Denton James W.}, How good are neural networks for causal forecasting, The
  Journal of Business Forecasting Methods and Systems 14~(2) (1995) 17--20.

\bibitem{hann_much_1996}
T.~H. Hann, E.~Steurer, Much ado about nothing? {Exchange} rate forecasting:
  {Neural} networks vs. linear models using monthly and weekly data,
  Neurocomputing 10~(4) (1996) 323--339.
\newblock \href {http://dx.doi.org/10.1016/0925-2312(95)00137-9}
  {\path{doi:10.1016/0925-2312(95)00137-9}}.

\bibitem{callen_neural_1996}
J.~L. Callen, C.~C.~Y. Kwan, P.~C.~Y. Yip, Y.~Yuan, Neural network forecasting
  of quarterly accounting earnings, International Journal of Forecasting 12~(4)
  (1996) 475--482.
\newblock \href {http://dx.doi.org/10.1016/S0169-2070(96)00706-6}
  {\path{doi:10.1016/S0169-2070(96)00706-6}}.

\bibitem{adhikari_combination_2013}
R.~Adhikari, R.~K.~Agrawal, A {Combination} of {Artificial} {Neural} {Network}
  and {Random} {Walk} {Models} for {Financial} {Time} {Series} {Forecasting},
  Neural Computing and Applications Accepted.
\newblock \href {http://dx.doi.org/10.1007/s00521-013-1386-y}
  {\path{doi:10.1007/s00521-013-1386-y}}.

\bibitem{omer_faruk_hybrid_2010}
D.~Omer~Faruk, A hybrid neural network and {ARIMA} model for water quality time
  series prediction, Engineering Applications of Artificial Intelligence 23~(4)
  (2010) 586--594.
\newblock \href {http://dx.doi.org/10.1016/j.engappai.2009.09.015}
  {\path{doi:10.1016/j.engappai.2009.09.015}}.

\bibitem{khashei_novel_2011}
M.~Khashei, M.~Bijari, A {Novel} {Hybridization} of {Artificial} {Neural}
  {Networks} and {ARIMA} {Models} for {Time} {Series} {Forecasting}, Appl. Soft
  Comput. 11~(2) (2011) 2664--2675.
\newblock \href {http://dx.doi.org/10.1016/j.asoc.2010.10.015}
  {\path{doi:10.1016/j.asoc.2010.10.015}}.

\bibitem{babu_moving-average_2014}
C.~N. Babu, B.~E. Reddy, A {Moving}-average {Filter} {Based} {Hybrid}
  {ARIMA}-{ANN} {Model} for {Forecasting} {Time} {Series} {Data}, Appl. Soft
  Comput. 23 (2014) 27--38.
\newblock \href {http://dx.doi.org/10.1016/j.asoc.2014.05.028}
  {\path{doi:10.1016/j.asoc.2014.05.028}}.

\bibitem{huang_empirical_1998}
N.~E. Huang, Z.~Shen, S.~R. Long, M.~C. Wu, H.~H. Shih, Q.~Zheng, N.-C. Yen,
  C.~C. Tung, H.~H. Liu, The empirical mode decomposition and the {Hilbert}
  spectrum for nonlinear and non-stationary time series analysis, Proceedings
  of the Royal Society of London A: Mathematical, Physical and Engineering
  Sciences 454~(1971) (1998) 903--995.

\bibitem{buyuksahin_siu_2018}
U.~C. Buyuksahin, S.~Ertekin, {Time} {Series} {Forecasting} {Using} {Empirical}
  {Mode} {Decomposition} and {Hybrid} {Method}, 26th Signal Processing and
  Communication Application Conference (SIU), IEEE, May 2018, Izmir Turkey.

\bibitem{wang_forecasting_2014}
J.~Wang, W.~Zhang, Y.~Li, J.~Wang, Z.~Dang, Forecasting wind speed using
  empirical mode decomposition and {Elman} neural network, Applied Soft
  Computing 23 (2014) 452--459.
\newblock \href {http://dx.doi.org/10.1016/j.asoc.2014.06.027}
  {\path{doi:10.1016/j.asoc.2014.06.027}}.

\bibitem{hibon_combine_2005}
E.~T. Hibon~M., To {combine} or not to {combine:} {selecting} {among}
  {forecasts} and {their} {combinations}. {International Journal of
  Forecasting}, International Journal of Forecasting 21 (2005) 15--24.

\bibitem{taskaya-temizel_comparative_2005}
T.~Taskaya-Temizel, M.~C. Casey, A comparative study of autoregressive neural
  network hybrids, Neural Networks: The Official Journal of the International
  Neural Network Society 18~(5-6) (2005) 781--789.
\newblock \href {http://dx.doi.org/10.1016/j.neunet.2005.06.003}
  {\path{doi:10.1016/j.neunet.2005.06.003}}.

\bibitem{intraday_epias}
Intraday {EPIAS},
  \url{https://www.epias.com.tr/en/intra-day-market/introduction}, 2018,
  (accessed 2018-05-14).

\bibitem{moosa_unbeatable_2014}
I.~Moosa, K.~Burns, The unbeatable random walk in exchange rate forecasting:
  {Reality} or myth?, Journal of Macroeconomics 40 (2014) 69--81.
\newblock \href {http://dx.doi.org/10.1016/j.jmacro.2014.03.003}
  {\path{doi:10.1016/j.jmacro.2014.03.003}}.

\bibitem{huang_applications_2003}
N.~E. Huang, M.-L. Wu, W.~Qu, S.~R. Long, S.~S.~P. Shen, Applications of
  {Hilbert}–{Huang} transform to non-stationary financial time series
  analysis, Applied Stochastic Models in Business and Industry 19~(3) (2003)
  245--268.
\newblock \href {http://dx.doi.org/10.1002/asmb.501}
  {\path{doi:10.1002/asmb.501}}.

\end{thebibliography}

\end{document}